\documentclass[runningheads]{llncs}

\usepackage[mobile]{accv}

\usepackage{accvabbrv}

\usepackage{graphicx}
\usepackage{booktabs}
\usepackage{amsmath}

\usepackage[pagebackref,breaklinks,colorlinks,citecolor=accvblue]{hyperref}

\usepackage{orcidlink}

\newcommand{\mycirc}[1][black]{\Large\textcolor{#1}{\ensuremath\bullet}}

\begin{document}

\title{HYPNOS : Highly Precise Foreground-focused Diffusion Finetuning for Inanimate Objects}

\titlerunning{HYPNOS}

\author{Oliverio Theophilus Nathanael\inst{1}\orcidlink{0009-0005-1206-9579} \and
Jonathan Samuel Lumentut\inst{1}\orcidlink{0000-0001-5146-8648} \and
Nicholas Hans Muliawan\inst{1}\orcidlink{0009-0000-1272-7272} \and
Edbert Valencio Angky\inst{1}\orcidlink{0009-0003-3065-5040} \and
Felix Indra Kurniadi\inst{2}\orcidlink{0000-0002-0850-601X} \and
Alfi Yusrotis Zakiyyah\inst{1}\orcidlink{0000-0002-6051-4279} \and
Jeklin Harefa\inst{1}\orcidlink{0000-0002-9486-6432}}

\authorrunning{O. T.~Nathanael et al.}

\institute{Universitas Bina Nusantara, Jakarta, Indonesia \and Università di Pisa, Pisa, Italy}

\maketitle

\begin{abstract}

    In recent years, personalized diffusion-based text-to-image generative tasks have been a hot topic in computer vision studies. A robust diffusion model is determined by its ability to perform near-perfect reconstruction of certain product outcomes given few related input samples. Unfortunately, the current prominent diffusion-based finetuning technique falls short in maintaining the foreground object consistency while being constrained to produce diverse backgrounds in the image outcome. In the worst scenario,  the overfitting issue may occur, meaning that the foreground object is less controllable due to the condition above, for example, the input prompt information is transferred ambiguously to both foreground and background regions, instead of the supposed background region only. To tackle the issues above, we proposed Hypnos, a highly precise foreground-focused diffusion finetuning technique. On the image level, this strategy works best for inanimate object generation tasks, and to do so, Hypnos implements two main approaches, namely: (i) a content-centric prompting strategy and (ii) the utilization of our additional foreground-focused discriminative module. The utilized module is connected with the diffusion model and finetuned with our proposed set of supervision mechanism. Combining the strategies above yielded to the foreground-background disentanglement capability of the diffusion model. Our experimental results showed that the proposed strategy gave a more robust performance and visually pleasing results compared to the former technique. For better elaborations, we also provided extensive studies to assess the fruitful outcomes above, which reveal how personalization behaves in regard to several training conditions.

\end{abstract}

\begin{figure}[t]
    \centering
      \begin{subfigure}{0.45\linewidth}
        \centering
        \includegraphics[width=.8\linewidth]{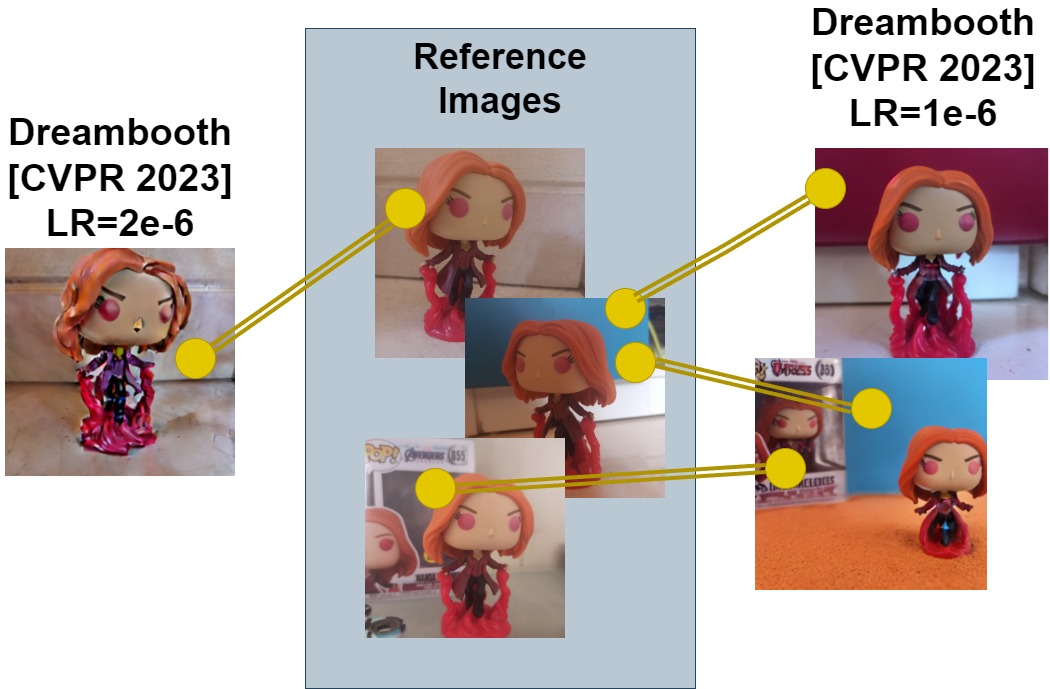}
        \caption{Entangled Foreground-Background}
        \label{fig:Dreambooth_Intro}
      \end{subfigure}
      \begin{subfigure}{0.47\linewidth}
        \centering
        \includegraphics[width=1\linewidth]{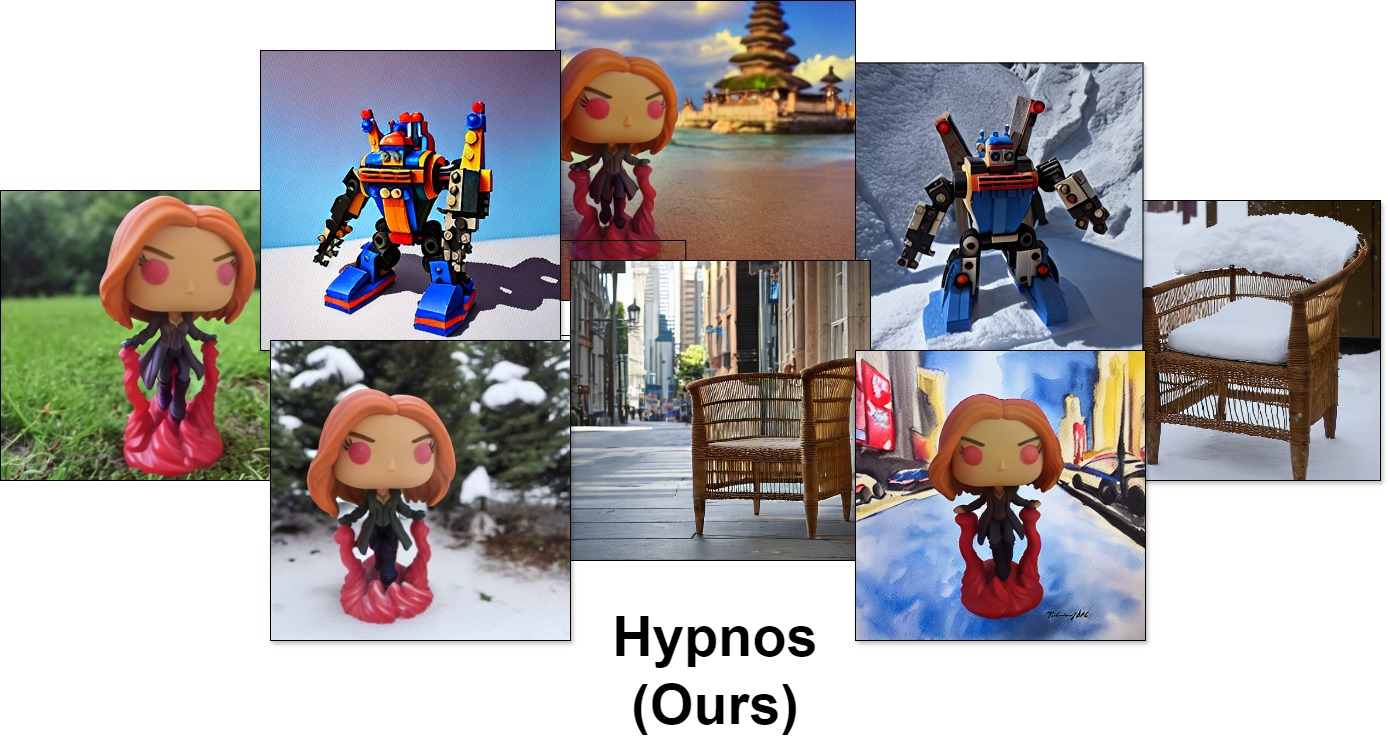}
        \caption{Hypnos generated images}
        \label{fig:Hypnos_Intro}
      \end{subfigure}
    \caption{The current (a) widely used method~\cite{1-Dreambooth} is prone to subject-scene entanglement and overfitting as shown by the yellow lines. Our proposed method visually (b) shows a realistic disentanglement effect between subject and scene at the foreground and background, respectively.}
    \label{fig:introduction}
\end{figure}

\section{Introduction}
\label{sec:intro}
Personalization for text-to-image (T2I) Diffusion Model has been a hot field of study and rapidly popularized  among researchers in computer vision studies as well as  practitioners and even hobbyist. 
Personalization is the key for generative models to be widely used as a tool for various use cases.
This personalization in the Diffusion Model has the same spirit as grounding Large Language Models (LLMs), where it opens the possibility of being reliably incorporated in diverse real-world environments. 
With reliable personalization in the Diffusion Model, it opens several potentials for recreational purposes and even in the business world, such as creating product advertisements and campaign photos, which mostly highlight the foreground objects (\textit{foreground-focused objective}).

One particular distinctive family of these techniques, namely Dreambooth~\cite{1-Dreambooth}, enables the ability to finetune Diffusion models on a specific image without losing prior knowledge. 
At the moment Dreambooth is one of the most popular personalization techniques because of its efficiency and minimum requirements of input samples (3-5 reference images) to be able to synthesize great quality, with high  similarities to the reference input. 

Despite that, there is still room for improvement, as the current method is still prone to structure and color distortions, which are mostly caused by the coarse attempt of the model to blend the object in the foreground and the scene in the background~\cite{1-Dreambooth}. 
Visually, there is an entanglement of background and foreground semantic information as shown in~\cref{fig:Dreambooth_Intro} (content relations are highlighted with yellow lines). 
To overcome this issue, one might increase the learning rate of~\cite{1-Dreambooth} to enforce more consistency to the reference images.
Unfortunately, such approach causes another problem to arise~\cite{1-Dreambooth}, which is the overfitting issue. 

To our observation, the work of~\cite{1-Dreambooth} tends to produce the entangled version of both foreground and background outputs from the given foreground reference input (\cref{fig:Dreambooth_Intro}).
Adjusting the learning rate to the higher (or lower) values in~\cite{1-Dreambooth} finetuning affects the variance of both foreground and background information.
This can be perceived as an entangled foreground-background issue in the T2I task, which is improper for the foreground-focused objective mentioned above.

In our pursuit of creating a reliable T2I finetuning technique, we introduce HYPNOS, a highly precise foreground-focused diffusion finetuning approach for inanimate objects.
We propose several strategies that work synergically to disentangle foreground-background information. 
These additions are proven to produce visually consistent and pleasing images as seen on~\cref{fig:Hypnos_Intro}. 
In detail, our strategies comprises of: straightforward image dataset augmentation, explicit foreground-background prompts procedures, and sets of new supervision mechanisms to quantify the foreground deviation from the reference images.  
Importantly, those method are implemented while still maintaining fast finetuning time and low latency architecture. 
We also conducted an extensive evaluation of both qualitative and quantitative analysis to further support our studies. 
These evaluations are also used to explain some distinct behaviors of Hypnos compared to Dreambooth and Textual Inversion techniques.
Our main contributions can be summarized as follows:
\begin{itemize}
    \item[$\bullet$] We propose a novel technique with competitive result by leveraging content-centric augmentation and new sets of supervision mechanisms. This improvement reliably ensures foreground-background disentanglement, significantly lowers noise, and enables semantic level tuning for the T2I task.
    \item[$\bullet$] We introduce prompt-invariant and prompt-varying metrics as a new approach to quantitatively assess the Dreambooth family technique and how to interpret the results. We show that these new metrics can be used to gain a better insight of the finetuned model.
    \item[$\bullet$] We conduct analysis and experiments on the introduced hyperparameters to elaborate further on their effects and how to tune them properly.
\end{itemize}

\section{Related Works}

\subsection{Vision-based Generative Model}
Diffusion Model is a groundbreaking image generation method that leverages thermodynamics based process \cite{31-sohldickstein2015deepunsupervisedlearningusing}. 
It was then further popularized by the introduction of the Denoising Diffusion Probabilistic Model \cite{3-DDPM}. 
It is a probabilistic model that is capable of modeling a complex distribution using a denoising process. 
Denoising Diffusion Implicit Model later eliminated the markovian assumption which speeds up the denoising process \cite{4-DDIM}. 
Other popular improvement includes the adaptation of Cosine scheduling on the noise diffusion process to ensure the image is not too rapidly destroyed throughout the diffusion process \cite{5-ImprovedDDPM}.
Other physics-inspired generative models include Poisson Flow Generative Models (PFGM), which is inspired by the physics of Electrodynamics \cite{6-PFGM}.
Its improvement, PFGM++ \cite{7-PFGM++}, unified PFGM Model with Diffusion Model.

The current trend involves the strategy of text-prompt-based input to guide the denoising process, hence called as T2I diffusion model.
CLIP, a transformer-based model that aligns text and image embedding \cite{8-radford2021learning}, is introduced to satisfy such task.
By utilizing CLIP, The work of GLIDE \cite{9-GLIDE} showed to produce great quality results.
Recent text transformer-based works, e.g. BERT \cite{10-BERT} and T5 \cite{11-T5}, also showed great performance in running the image generative task \cite{12-saharia2022photorealistic}.

Diffusion process recently involved the denoising procedure on Variational Autoencoder's \cite{14-Kingma_2019} latent space to make the model resource-effective.
This method is commonly known as the Latent Diffusion Model \cite{13-rombach2022highresolution}.
The following are among the popular latent-based works: Stable Diffusion \cite{13-rombach2022highresolution}, Stable Diffusion XL \cite{15-podell2023sdxl}, Dall-E 3 \cite{16-shi2020improving}, and the recent video-based (temporal) diffusion generative task\cite{17-blattmann2023stable}.

\subsection{Personalization}
As explained on \cref{sec:intro} personalization aims to enable T2I models to be able to generate a specific object based on some reference images.
Some of the methods include textual inversion \cite{2-TextualInversion} and controlnets \cite{18-zhang2023adding}. 
Among many personalization techniques, the one that arguably stands out the most is Dreambooth \cite{1-Dreambooth}. 
Dreambooth finetunes a diffusion model using 3-5 reference images and some prior class images to prevent the loss of prior ability to produce diverse images on that particular class.
Dreambooth also can be trained relatively fast.

In recent years, numerous works have been proposed to better fit Dreambooth in some use cases.  
LoRA \cite{19-hu2021lora} speeds up the finetuning process at the expense of image quality.
Google's HyperDreamBooth \cite{20-ruiz2023hyperdreambooth} focuses on fast personalization for human faces using Hypernetworks to adjust main network weights \cite{21-chauhan2023brief}; further accelerated by LoRA.
DreamCom~\cite{22-lu2024dreamcom} had shown remarkable results on the image inpainting task by reusing some good generated images to retrain the model.
Other variations of Dreambooth include the procedure of learning other concepts besides the object appearance itself \cite{23-motamed2023lego}.
In this work, we opt to scope the usage of Dreambooth~\cite{1-Dreambooth} (comprises of Stable Diffusion \cite{13-rombach2022highresolution}) as the baseline for our foreground-focused diffusion finetuning task on inanimate objects.

\section{Methodology}
\begin{figure}[t]
    \centering
    \includegraphics[width=1\linewidth]{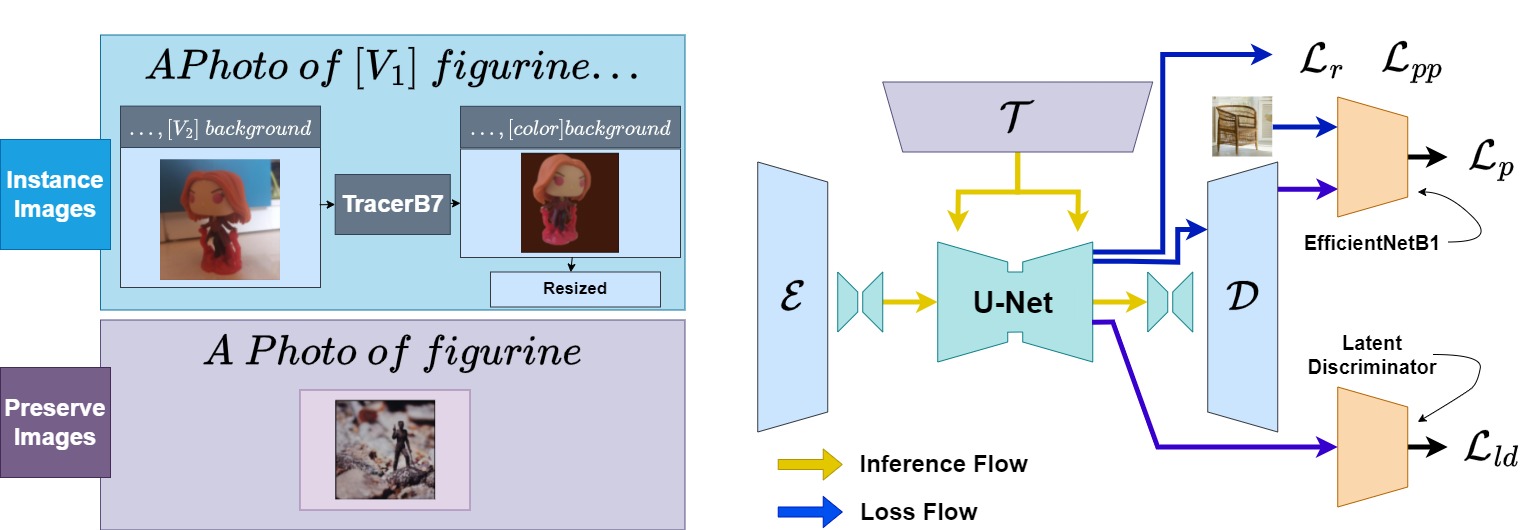}
    \caption{We proposed to apply image augmentation and prompt engineering to the instance dataset, we also apply sets of losses as shown by the blue arrows}
    \label{fig:Mainflow}
\end{figure}

The proposed method is crafted to accommodate Dreambooth in terms of producing a more consistent foreground subject.
Hence, we utilized the same backbone model, but with additional proposed functionalities to better disentangle the foreground-background information.
As seen on~\cref{fig:Mainflow}, there are two main part of the process. 
The~\textit{first one} is the datasets refinement procedure that undergoes our specific image augmentation and prompt engineering functionalities (left side of~\cref{fig:Mainflow}). 
The~\textit{second part} is the Diffusion model itself that is trained on our diverse supervision mechanisms (right side of~\cref{fig:Mainflow}).

\subsection{Dataset Refinement}
Similar to Dreambooth, Hypnos is a finetuning technique that only need 3-5 image samples. 
We limit the scope of the generation task to only for inanimate objects. 
Our images that are used within this work are taken from personal images and Unsplash (https://unsplash.com/). 

\subsection{Text to Image Model}
To generate the image we use a stable diffusion v1.5 model~\cite{13-rombach2022highresolution}. 
The model itself comprises of several networks that work synergically. 
These models usually utilize U-Net as the network to model the denoising process. 

Stable Diffusion is a latent diffusion model which means that the diffusion and denoising process happens on a latent representation of an image.
Furthermore, as a Text to Image model, it also incorporates CLIP embeddings into the U-Net model to guide the generated image based on a specific prompt. 
Following the common procedure, the images in our method are encoded by the Variational Autoencoders (VAE), while the text prompts are processed by the CLIP method.

In terms of finetuning, there are several approaches to run such task using the Dreambooth method. 
In this work, we opt to freeze the weight of the VAE, meaning that Hypnos only optimize both the denoising model and the text encoder. 
We also preserve the denoising procedure from the Stable Diffusion which maintain the lower latency characteristic~\cite{13-rombach2022highresolution}.

\subsection{Image Augmentation and Prompt Engineering Functionalities}
\label{sec:image_aug}
Based on the original Dreambooth paper there are two sets of dataset to be extracted for supervision (\cref{fig:Mainflow} (left)). 
The~\textit{first one} ({\color{violet}purple region} of~\cref{fig:Mainflow} (left)) is the self generated class images paired with standard prompt such as "\underline{a photo of [C]}" where [C] is the class name of the subject. 
The~\textit{second set} ({\color{blue}blue region} of~\cref{fig:Mainflow} (left)) is the specific provided subject images and traditionally labeled as "\underline{a photo of [V] [C]}"  where [V] is a rare word that acts as a name to tell our object apart from others.
Both extracted datasets are utilized in supervision to prevent the lost of the prior general understanding of that particular class.

Unfortunately, doing the above's strategy in Dreambooth \cite{1-Dreambooth} invokes the entanglement effect in both foreground and background, which is unfavorable for foreground-focused objective. 
To avoid this, we propose a content-centric augmentation that explicitly introduces background-foreground disentanglement. 
We then utilized TracerB7 \cite{24-lee2022tracer, 25-Qin_2020} to randomly replace the background with monotone colors.
We empirically set the augmentation proportion to 0.66 to avoid losing the ability to blend the subject with the background. 
To further introduce the foreground diversity, we also resized a small portion of the sampled dataset spatially.
Moreover, we provided~\textit{prompting adjustment strategy} to allow the model to recognize the foreground and background via first and second clauses, respectively, as seen on \cref{fig:Mainflow}
To describe the original image's background, it is impractical to use a particular color name as used in the background changed images; hence, we opted to utilize another rare word as a placeholder.

\subsection{Supervision Mechanisms}
\label{sec:loss}
One of our proposed method includes specialized losses to make sure that the model will preserve the foreground structure and color while still letting the background to exhibit a high variance. 
Our proposed method is a combination of 4 different losses which are reconstruction losses, prior preservation loss, perceptual loss, and latent discriminator loss as shown with the blue arrows on \cref{fig:Mainflow}, which is written as \cref{eq:loss}:

\begin{equation}
  \mathcal{L} = \lambda_r\mathcal{L}_r + \lambda_{pp}\mathcal{L}_{pp} + \lambda_p\mathcal{L}_p + \lambda_{ld}\mathcal{L}_{ld}.
  \label{eq:loss}
\end{equation}

Based on experiments it is generally best to set \(\lambda_r=1\), \(\lambda_{pp}=1\), \(\lambda_p=0.003\), \(\lambda_{ld}=0.5\). 
Note that these weights itself are hyperparameters, hence it can be adjusted based on the intended model output.

\subsubsection{Reconstruction Loss ($\mathcal{L}_{r}$)}
Traditionally this loss is simply the mean squared error of the predicted noise of the instance latent images which denotes by \(z_i\) at timestep \(t\) where \(t \sim \mathcal{U}(1, 1000)\). 
This loss ensures the generated image is as similar as possible to the instance image. 
Though mean squared error is a widely known loss for reliably quantifying deviation from a target value, our observation suggests that exponential-growth-base loss function is preferable than quadratic-growth-base one.
This is done as measuring the deviation from a noise on a latent space is focused on the data with normal distribution.
One intuitive alternative is to use the inverse of gaussian distribution as shown on \cref{eq:invertedgauss} as it is based of the VAE overall distribution itself. 
\begin{equation}
  \mathcal{L}_r = \sigma\sqrt{2\pi}(\mathbb{E}_{z_i, t, c, \epsilon}e^{\|{\epsilon_\theta(z_i, t, c)-\epsilon}\|^2_2/{2\sigma^2}}-1)
  \label{eq:invertedgauss}
\end{equation}
Based on our experiment, this is very effective for getting rid of the noise artefact that is often seen in a higher learning rate of Dreambooth method (details on~\textit{Supplementary Material}).
By adjusting the $\sigma$ it is possible to tune on how steep the loss is. 
In detail, the L2 loss can be re-approximated by tweaking $\sigma$, particularly by minimizing the least squared regression of both function for bound $\pm a$ as shown on \cref{eq:regression}.
\begin{equation}
  \min_\sigma\int^a_0(x^2- \sigma\sqrt{2\pi}(e^{x^2/{2\sigma^2}}-1))^2dx
  \label{eq:regression}
\end{equation}

By solving \cref{eq:regression} for $a$ equals to 1, where reconstruction loss are very unlikely to exceed 1, it is empirically found that $\sigma=1.382$ is the closest to L2 loss, with standard deviation larger than that, meaning that the loss is flatter than L2 loss.

\subsubsection{Prior Preservation Loss ($\mathcal{L}_{pp}$)}

While Reconstruction Loss computes the deviation between the generated image with respect to the subject latent images, Prior Perceptual loss works with respect to the class latent images  which denotes by $z_p$ as written in \cref{eq:priorpreserve}:
\begin{equation}
  \mathcal{L}_{pp} = \mathbb{E}_{z_p, t, c, \epsilon}e\|{\epsilon_\theta(z_p, t, c)-\epsilon}\|^2_2.
  \label{eq:priorpreserve}
\end{equation}
In this work, we opt to preserve the original MSE loss on this prior preservation function, as this allows the model to focus its learning process on the instance images rather than the class images.

\subsubsection{Perceptual Loss ($\mathcal{L}_{p}$)}

Perceptual Loss is widely used for style transfer task. The advantage of using this loss is that it can semantically quantify the deviation of the generated image. 
We utilized EfficientNetB1 \cite{26-tan2020efficientnet} (a classifier network) as the encoder and take the L2 loss across several of its activation.  
To preserve low-level semantic information, we gave larger weight to the activations of its shallow layers
Furthermore, using classifier network to guide a diffusion model is a common practice. 
One justification includes the use of classifier output to aid an ambiguous text to produce an image that is better represent the intended picture \cite{27-schwartz2023discriminative}.

Since classifier networks process decoded denoised images, thus, to be able to apply perceptual loss, it is required to denoise the images straight to $z_0$ and then decode the image beforehand. 
The resulting equation is described on \cref{eq:perceptual}:
\begin{equation}
  \mathcal{L}_p = \mathbb{E}_{z_i, t, c, x}\|{C_\theta(x)-C_\theta(D_\theta(z_0))}\|^2_2,
  \label{eq:perceptual}
\end{equation}
where $C_{\theta}$ is a classifier model, $D$ denotes the VAE Decoder, $x$ denotes the reference image, and $z_0$ as the denoised latent.

Based on our observation, to avoid overfitting, there are two approaches that can be done, \textit{the first one} is to apply Latent Discriminator that balances the loss and \textit{second one} is by limiting its influence on the loss, rather than decreasing the loss weight. 
The latter can be implemented by limiting the number of steps that are influenced by Perceptual Loss. 
Intuitively, this is done to avoid local minima of the original Dreambooth loss that exhibit a high perceptual dissimilarity as described on \cref{fig:Dreambooth_Intro}. 
It is also important to avoid a perfect perceptual match. 
Therefore, Perceptual Loss should guide optimization for only a certain number of steps, with the remaining steps optimized without it.
To do so, we introduced a new hyperparameter $s_p$ that directly control the perceptual similarity of the generated image by limiting the maximum step influenced by Perceptual loss.
We set $s_p=500$ out of 800 steps of total training cycle on the re-defined Perceptual loss $\mathcal{L}_{p}$ in \cref{eq:perceptualnew}.

\begin{equation}
  \mathcal{L}_p = 
  \begin{cases}
      \mathbb{E}_{z_i, t, c, x}\|{C_\theta(x)-C_\theta(D_\theta(z_0))}\|^2_2, &\text{if } s \leq s_p \\
      0, & \text{otherwise}
  \end{cases}
  \label{eq:perceptualnew}
\end{equation}

\begin{figure}[t]
  \centering
  \begin{subfigure}{0.55\linewidth}
    \includegraphics[width=1\linewidth]{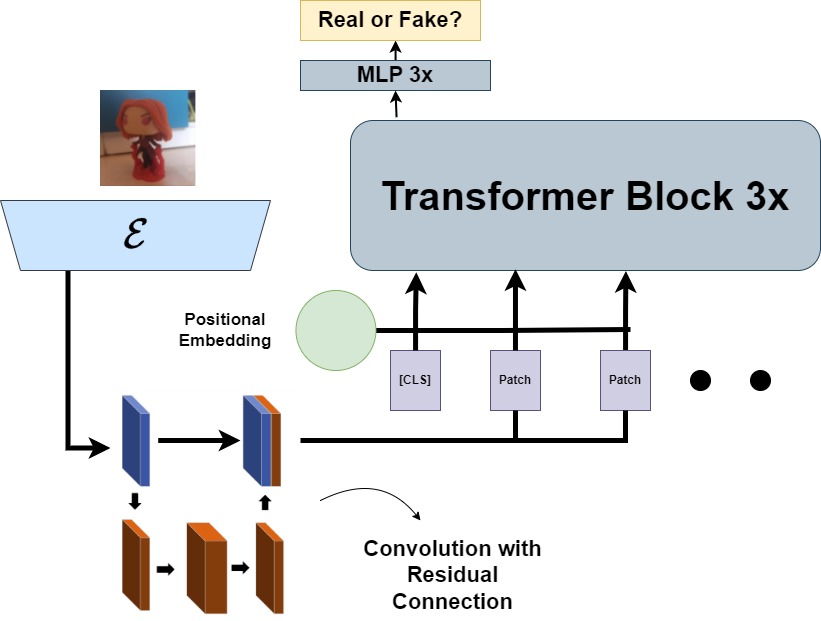}
    \caption{Latent Discriminator Model}
    \label{fig:ld_model}
  \end{subfigure}
  \hfill
  \begin{subfigure}{0.4\linewidth}
    \includegraphics[width=1\linewidth]{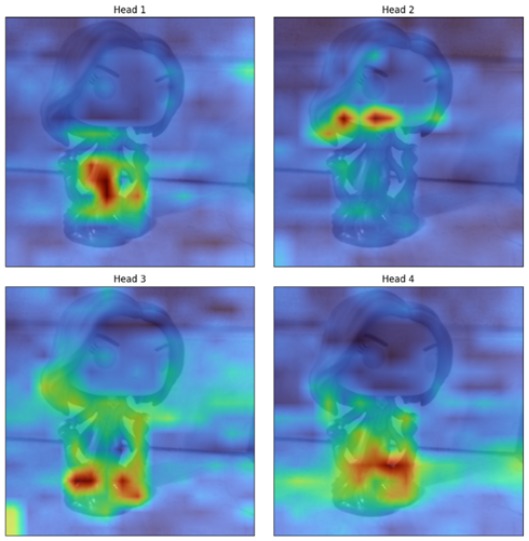}
    \caption{Latent Discriminator Attention Head Visualization}
    \label{fig:attention}
  \end{subfigure}
  \caption{Latent Discriminator is designed to be lightweight and has low latency. Despite the small size, the discriminator successfully identify important feature as shown by the attention head visualization }
  \label{fig:ld}
\end{figure}

\subsubsection{Latent Discriminator Loss ($\mathcal{L}_{ld}$)}
Similar to perceptual loss, this loss aims to quantify the foreground deviation semantically to preserve the foreground information. 
Rather than computing the loss of the decoded image, this loss computes the difference straight from the denoised latent space.
To achieve this, we make a discriminator trained on the latent representation of the instance image and the monotone colored background version as the real images while fake images comprises of class images and altered instance images such as removed foreground and negative colored foreground (details on~\textit{supplementary material}). 
The discriminator itself is a 3 layer vision transformer \cite{28-dosovitskiy2021image} with additional convolutional layers to ensure efficient semantic information extraction (visualized in \cref{fig:ld_model}).  

Similar to Adversarial Diffusion Distillation \cite{29-sauer2023adversarial} that applied adversarial learning on diffusion model, our latent discriminator is also provided with the flexibility to compete with main model. 
It is written in \cref{eq:ld} with $LD_\theta(.)$ and $z_0$ denote the Latent Discriminator model and denoised latent image, respectively:
\begin{equation}
  \mathcal{L}_{ld} = \mathbb{E}_{z_i, t}\|{1-LD_\theta(z_0)}\|^2_2.
  \label{eq:ld}
\end{equation}

The initial intention of this loss ($\mathcal{L}_{ld}$) is to provide the diffusion model with an explicit semantic similarity loss referenced to all instance images.
By this we can utilize all image references rather than just one sample as in the other Losses ($\mathcal{L}_r$, $\mathcal{L}_{pp}$, $\mathcal{L}_p$).
Thus, adversarial learning can be viewed as an effort for the discriminator model only to adapt with the everchanging image quality of the generated image. 
We design our discriminator with light architecture and train it initially with 600 steps, and then followed by the combined training.

The implementated architecture for Latent Discriminator is shown on \cref{fig:ld_model}. 
As seen on \cref{fig:attention}, this simple architecture  is enough to detect important features of the object.

\subsection{Evaluation Metrics}
\label{sec:metrics}
Quantifying Dreambooth method family is one of the most tricky task to handle there are at least two main arguments to support this claim. The first one is that Dreambooth family falls into few-shot learning category which by definition has a scarce amount of ground truth, hence it is not suitable to represent the approximate intended data distribution, rather it is expected to be able extrapolate from that given data distribution. This imply that the measure of how good the extrapolation result vary a lot across preferences and use cases. Second, ground truth incorporating diverse prompt are non-existent, unlike what seen in classic Diffusion model task or even controlnet model \cite{18-zhang2023adding}. 
This leads to some evaluation blind spot on how well model can adapt to diverse prompts and how does the model adapt to the prompt and what are the impact to the object. 

\subsubsection{Prompt Invariant and Prompt Varying Evaluation}
In this work, we proposed a modification to the existing metric evaluation to open new perspective and used it alongside the existing method. 
This is reasonable since Dreambooth family techniques usually use standard prompt to generate all the images which we called as~\textit{Prompt Invariant Evaluation}. 
To better accomodate the current evaluation approach, we propose that it is also important to assess the adaptibility of the resulting model on diverse prompt, which we coined as~\textit{Prompt Varying Evaluation}.

\begin{figure}[t]
    \centering
    \includegraphics[width=.7\linewidth]{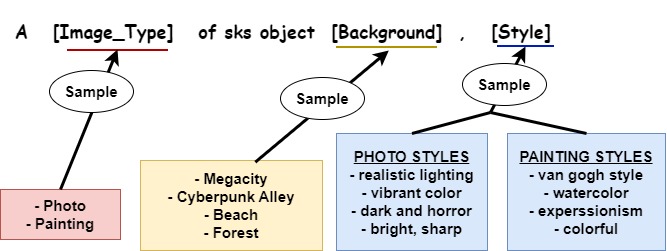}
    \caption{Prompts for Varying Prompt Evaluation are sampled from the combination of predefined list for image type, background, and style}
    \label{fig:proptgen}
\end{figure}

To minimize bias that emerges while handcrafting every single prompt, we proposed a solution that samples parts of a generic image generation prompt structure as seen on \cref{fig:proptgen}. 
There are three main considerations, the first one is the~\textit{\textbf{image type}} such as photo or painting, the second part is the~\textit{\textbf{background information}}. 
In this context, background information is not only limited on the background scene but also the image composition other than the main object itself. 
The third one is the~\textit{\textbf{styles}}; style descriptions are crucial to generate high quality image in smaller diffusion models as they require longer prompt compared to larger model such as SDXL\cite{15-podell2023sdxl}. 
Based on our understanding, style descriptions may vary across image types, hence its description is acquired from the previously sampled image type. 

\subsubsection{Dreambooth Metrics}
As perceived from the Dreambooth work, their finetuning technique is a unique task that is inadequate to be measured using classic image generation metrics because of its nature that have limited reference image as explained earlier. 
Following their scheme, we use DINO\cite{30-caron2021emerging}, CLIP-I, and CLIP-T to perform the baseline quantitative measurement.

\subsubsection{Image Quality and Perceptual Metrics}
Apart from the metrics above, image generation tasks are also evaluated using metrics suvh as FID\cite{32-FID}, LPIPS\cite{33-LPIPS}, SSIM, and PSNR
These metrics are rarely used to evaluate the Dreambooth method family as it is prone to biased score. 
However, we still opt to include them as they still can provide additional insight on the generated image quality.
\section{Result and Discussion}

\subsection{Evaluation}
\label{sec:result}
As mentioned on \cref{sec:metrics}, the evaluation of Dreambooth method family can vary across preference and use cases. 
Therefore, we conduct both qualitative and quantitative analysis of the results. 
As shown later on this section, solely analyzing quantitative metrics on this particular task has a potential to be misleading, hence it is advised to interpret the quantitative assessment with a grain of salt. 
We resolve the issues above by employing the procedure of always combining quantitative and qualitative analysis throughout the evaluation process.

\begin{table}[tb]
  \caption{\textbf{\textit{Prompt Invariant}} quantitavie metrics evaluated on 3 datasets, Funko figurine ({\mycirc[red]}), Rattan chair ({\mycirc[green]}), and Lego Robot ({\mycirc[blue]}).
  }
  \label{tab:invariant}
  \centering
  \scalebox{0.9}{
  \begin{tabular}{@{}l@{\hspace{4pt}}ll@{\hspace{10pt}}l@{\hspace{10pt}}l@{\hspace{10pt}}l@{\hspace{10pt}}l@{\hspace{10pt}}l@{\hspace{10pt}}l@{}}
    \toprule
    Method &   &DINO & CLIP-I & CLIP-T & FID& SSIM& PSNR&LPIPS\\
    \midrule
    Hypnos (Ours) &  \mycirc[red]&\bf{0.7851}& \bf{0.8635}
 & 0.0094
 & 3,6032
& 0.5974
& 11.8504
&0.3850
\\
  &  \mycirc[green] &\bf{0.6502}& \bf{0.8015}&0.0067
 & \bf{2.3840}& \bf{0.2225}& \bf{9.4634}&\bf{0.4166}\\
  &  \mycirc[blue]&0.6589& 0.8369 &0.0183 & \bf{5.6330}& 0.3876& \bf{10.5387}&0.4624\\
 &  && & & & & &\\
    Dreambooth &  \mycirc[red]&0.6422
 &  0.7935
& 0.0183
 & \bf{2.9873}& \bf{0.6056}& \bf{12.2883}&\bf{0.3663}\\
  (LR=1e-6)&  \mycirc[green]&0.5012
 &  0.7404
&\bf{0.0549} & 13.1933
& 0.1645
& 9.0604
&0.4583
\\
  &  \mycirc[blue]&\bf{0.7130}&  \bf{0.8753}&\bf{0.0458} & 5.7367& 0.3429& 9.3005&0.4679\\
 &  && & & & & &\\

    Dreambooth &   \mycirc[red]&0.5311
&  0.7468& 0.0153
 & 14.7671
& 0.4781
& 11.4756
&0.4513
\\
 (LR=2e-6)&  \mycirc[green]&0.2647
& 0.4742&0.0224
 & 42.7634
& 0.1433
& 9.3789
&0.5128
\\
 &  \mycirc[blue]&0.5704& 0.8323&0.0175 & 14.2261& 0.3060& 9.3813&\bf{0.4622}\\
 &  && & & & & &\\
    Textual &   \mycirc[red]&0.4934
&  0.6469
& \bf{0.0417} & 12.2159
& 0.4565
& 9.9478
&0.4917
\\
 Inversion&  \mycirc[green]&0.4397
& 0.7134
&0.0308
 & 4.6512
& 0.2125
& 8.8142
&0.4785
\\
 &  \mycirc[blue]&0.3904& 0.6118&0.0312 & 6.4942& \bf{0.3929}& 9.6875&0.5160\\
 \bottomrule
  \end{tabular}
  }
\end{table}

\subsubsection{Prompt Invariant Evaluation} 
As seen on \cref{tab:invariant}, we evaluated each metrics on three subject samples which are a Funko figurine ({\mycirc[red]}), a rattan chair ({\mycirc[green]}), and a Lego robot ({\mycirc[blue]}). 
This approach can be observed on \cref{tab:invariant}.
We opt to experiment on these samples to minimize the bias effect while assessing the metrics. 
The prior method of Dreambooth utilized a small number of instance images that were insufficient to be referred to as the perfect target distribution.
This yields the metrics results that vary vastly across datasets, which we desire to avoid.
By this reasoning, we performed a comparison if and only if those methods were trained with the same instance images.

\begin{figure}[t]
    \centering
    \includegraphics[width=1\linewidth]{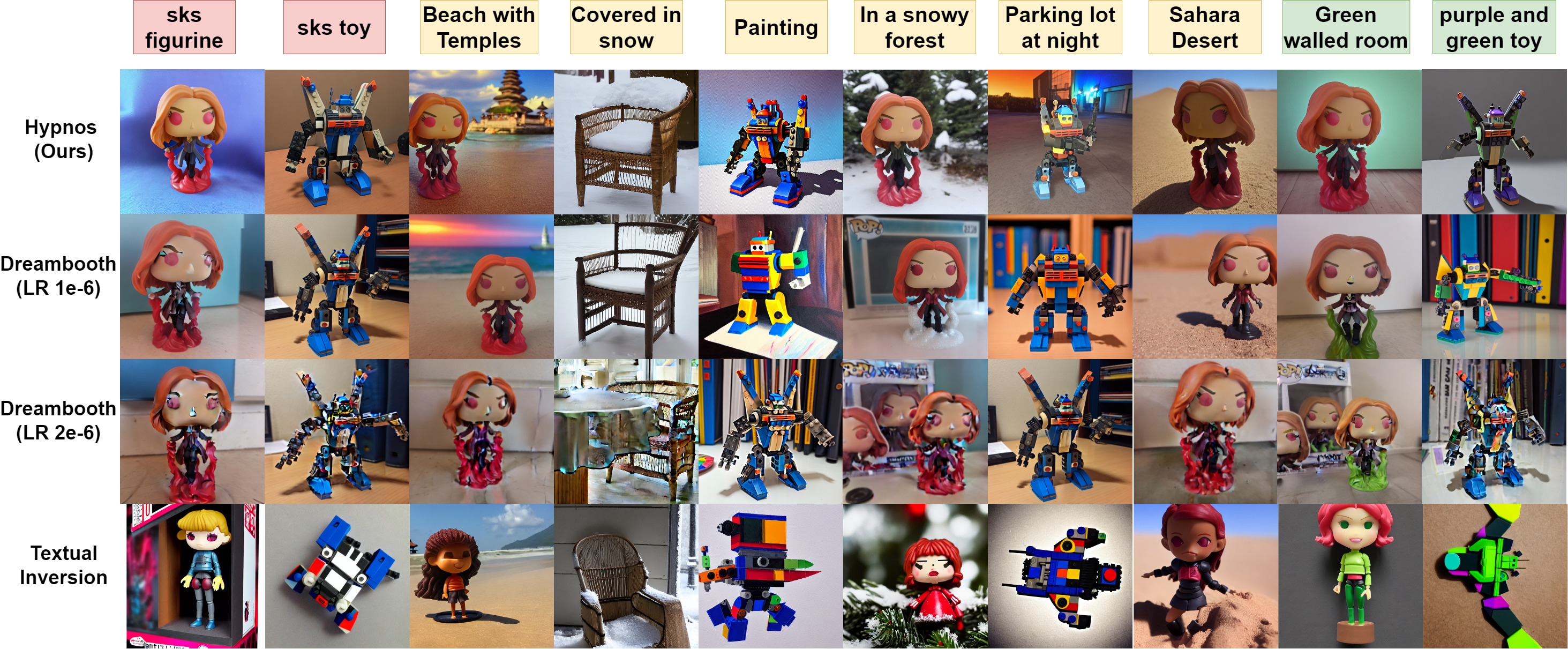}
    \caption{Image generation comparison, {\color{red}red} prompt denotes \textit{prompt invariant}, {\color{yellow}yellow} prompt denotes \textit{prompt varying}, {\color{green}green} prompt denotes specific prompting to analyze foreground-background disentanglement ability and highlight semantic leaking}
    \label{fig:Result}
\end{figure}

As shown in \cref{tab:invariant}, Hypnos is consistently superior in DINO and CLIP-I metrics.
In contrast, Dreambooth mainly dominates CLIP-T.
However, CLIP-T's metric is the least aligned with qualitative outcomes (refer to CLIP-T scores in \cref{tab:invariant} vs. results in \cref{fig:Result}).
We suspect that this is caused by the large domain gap between post-training CLIP text embedding and pre-trained CLIP image embedding, which, most of the time, each embedding deviates and becomes unreliable.
DINO and CLIP-I, on the other hand, are rather in line with what was observed qualitatively.
Visually, Hypnos exhibits minimum noise effect in the generated images while preserving most foreground information similar to the dataset.
Dreambooth, even with a lower learning rate case, produces high-noise results (\textit{first} and~\textit{second column} in \cref{fig:Result}) and tends to overfit on the instance image.
This is reasonable because both the image and prompt are what exactly the model learned in the training phase. 
In contrast, Hypnos introduced higher variance in the training background with its respective prompt, resulting in arbitrary background results when no background information is provided in the prompt input.
At the same time, Hypnos is still able to maintain structurally similar foreground outputs (\textit{first} and~\textit{second column} in \cref{fig:Result}).

Based on our observation, Dreambooth tended to yield high noise when the object or scene comprises of mostly homogeneous regions (e.g. figurine case).
On the Lego robot case, Dreambooth gave perfect generation results although lower learning rate is utilized and thus it produced higher scores in the metrics.
Nonetheless, the noise level becomes aggravated in the Lego case if a higher learning rate is employed.
Hypnos tackles the above problem as shown in the first row of \cref{fig:Result}.
In the case of longer prompts input ({\color{yellow}yellowish}  prompt in \cref{fig:Result}), unlike Dreambooth, Hypnos avoids the overfit and underfit effects, which yields the production of highly similar and high-quality foreground results.

Furthermore, as seen on \cref{fig:Result} (\textit{fourth row}), Textual Inversion produced a poor output in terms of similarity even with the standard prompt usage. 
From this point onward, Textual Inversion acted as a baseline and control to our metrics analysis as it is already widely known to underperform compared to the classic Dreambooth.

\cref{tab:invariant} also shows standard image generation metrics with the prompt invariant case.
The metrics provided useful insights, yet they are insufficient to determine model superiority.
Nevertheless, Hypnos still dominates most of the metrics.
In terms of FID, Hypnos consistently generated a perceptually similar image to the instance images. 
This is further supported by higher SSIM and PSNR scores (highlighted with~\textbf{bolds} in \cref{tab:invariant}; \textbf{bolds} indicate best scores).

\begin{table}[tb]
  \caption{\textbf{\textit{Prompt Varying}} quantitavie metrics evaluated on 3 datasets,  Funko figurine ({\mycirc[red]}), Rattan chair ({\mycirc[green]}), and Lego Robot ({\mycirc[blue]}).
  }
  \label{tab:varying}
  \centering
  \scalebox{0.9}{
  \begin{tabular}{@{}l@{\hspace{4pt}}ll@{\hspace{10pt}}l@{\hspace{10pt}}l@{\hspace{10pt}}l@{\hspace{10pt}}l@{\hspace{10pt}}l@{\hspace{10pt}}l@{}}
    \toprule
    Method &   &DINO & CLIP-I & CLIP-T & FID& SSIM& PSNR&LPIPS\\
    \midrule
    Hypnos (Ours) &  \mycirc[red]
&\bf{0.7070}& 0.7883
& 0.0200
& 11.0675
& 0.5139
& 10.9563
&0.4626
\\
  &  \mycirc[green]
&\bf{0.5461}& \bf{0.6572}&\bf{0.0326}& \bf{8.6402}& \bf{0.1797}& 8.8435
&0.5039
\\
  & \mycirc[blue]
&0.4920& 0.6462&0.0242& 22.5143& 0.2863& \bf{9.7863}&0.5392\\
 &  
&& & & & & &\\
    Dreambooth &  \mycirc[red]
&0.7050
&  0.7837
& 0.0224
& 6.5111
& 0.5453
& 10.7109
&0.4402
\\
  (LR=1e-6)&  \mycirc[green]
&0.4499
&  0.5814
&0.0173
& 11.2734
& 0.1687
& 8.1098
&0.5196
\\
  &  \mycirc[blue]
&0.4377&  0.6685&0.0286& 14.6446& 0.2887& 9.1872&0.5502\\
 &  
&& & & & & &\\
    Dreambooth &   \mycirc[red]
&0.6630
&  \bf{0.8028}& 0.0204
& \bf{5.1134}& \bf{0.5589}& \bf{11.6786}&\bf{0.4089}\\
 (LR=2e-6)&  \mycirc[green]
&0.4656
& 0.6424
&0.0179
& 17.1525
& 0.1650
& \bf{9.1083}&\bf{0.4583}\\
 &  \mycirc[blue]
&\bf{0.5826}& \bf{0.7704}&\bf{0.0336}& \bf{10.4226}& \bf{0.3325}& 9.7451&\bf{0.4830}\\
 &  
&& & & & & &\\
    Textual &   \mycirc[red]
&0.3131
&  0.4355
& \bf{0.0297}& 42.3451
& 0.3066
& 8.9635
&0.5942
\\
 Inversion&  \mycirc[green]
&0.3242
& 0.5427
&0.0224
& 18.9589
& 0.1273
& 7.8409
&0.5455
\\
 &  \mycirc[blue]&0.3132& 0.5051&0.0239& 25.5648& 0.2397& 8.6579&0.5935\\
    \bottomrule
  \end{tabular}
}
\end{table}

\subsubsection{Prompt Varying Evaluation}
\cref{tab:varying} shows the metrics score across varying prompts. 
Varying prompts is expected to force the models to produce diverse images.
This implies that the similarity index should decrease significantly. 
Those performances are observed in Hypnos when comparing the scores in \cref{tab:varying} and \cref{tab:invariant}. 
A noticeable change is that the higher learning rate Dreambooth dominates most metrics, suggesting high similarity.
This phenomenon can be explained by observing the resulting images as seen on the~\textit{third row} in \cref{fig:Result}. 
Qualitatively, it is apparent that Dreambooth (LR=2e-6) is overfitted, and the generated images fail to adapt to the prompt.
Thus, Dreambooth’s high scores are obtained from overfitting, which is inadequate.

On the other hand, Dreambooth, with a lower learning rate, seems to be able to better adapt to the prompt.
However, as mentioned earlier, they came with a downside effect on the foreground quality.
As seen on the~\textit{fifth} and~\textit{eighth column} of \cref{fig:Result}, it is capable of adapting to the prompt input, but its foreground object is highly altered. 
Hypnos, in contrast, exhibited a more consistent and robust foreground similarity. 

Another downside of Dreambooth is that it is often overfit on the background.
As seen on the~\textit{third column} of Dreambooth case of \cref{fig:Result}, the beach looks like a room, and on the~\textit{sixth column}, Dreambooth fails to produce a forest scene.
It instead overfits the dataset background (\textit{second row} of \cref{fig:Result}).
Another example is shown on the~\textit{fifth} and~\textit{seventh column} where the toy clearly always stands in front of books and fails to follow the prompt; this is because some of the training images are taken in front of the books.
These issues are effectively resolved on Hypnos-generated images as they are trained to have content-centric knowledge.

Hypnos also displayed a clear distinction between foreground and background.
This statement is justified by how it preserves the image color and adapts to the scene naturally.
On the other hand, Dreambooth showed the foreground-background entanglement effect, in which its phenomenon is observed on the~\textit{sixth} and~\textit{eighth column} of \cref{fig:Result}, where the base of the figurine's colors are affected by the scenes' color.

Furthermore, Hypnos' robust foreground structure preservation allows it to produce more imaginative images, such as snow on a rattan chair (\textit{first row fourth column} of \cref{fig:Result}).
In that case, both components are rarely seen together in the real world. 
Dreambooth, however, generated a very different chair and struggled with high-frequency features like a rattan chair (\textit{fourth column} of \cref{fig:Result}).
This explains why Hypnos dominates the metrics for Rattan Chair as seen on both tables (\cref{tab:invariant} and \cref{tab:varying}).

\subsubsection{Foreground-Background Disentanglement Analysis}
Hypnos also solved another limitation of Dreambooth~\cite{1-Dreambooth}, mentioned in their work, which is the color and theme of the scene that leaks information from the foreground object or vice versa (entangled).
As seen on \cref{fig:Result} with green prompts, Dreambooth produced a highly overfitted image with noises when prompted with green walled room. 
Rather than coloring the room wall green, it painted the base green instead, and this phenomenon is frequently observed on Dreambooth. 
This phenomenon is also clearly observed in textual inversion, where the clothes are green instead of the wall.  
By leveraging content-centric strategies, Hypnos, in contrast, can correctly make a green walled room background while still having accurate foreground color (refer to the second column from right in \cref{fig:Result}).

Besides excelling in background modification, Hypnos also excels in foreground modification.
This is observed to be challenging for Dreambooth, especially for the more complex objects where they often end up overfitting even with a lower learning rate (first column from the right of \cref{fig:Result}). 
Dreambooth structural consistency dropped with longer prompt and having difficulty adapting to the prompt whereas Hypnos successfully altered the object color to purple and green without altering much of the structure. 
This again proves the robustness of Hypnos in generating inanimate objects in various scenes and prompts.

\subsection{Ablation Studies}
\label{sec:ablation}
\subsubsection{Perceptual Loss}
We evaluate Hypnos on two configurations to analyze the effect of Perceptual loss.
The First one is when we exclude any limit to the maximum step influenced by Perceptual Loss. 
In this case, it tends to highly overfit, such as the same lighting and pose information directly pasted from the original image. 
We also observed the emergence of background overfitting because of this uninformed subject-scene loss.
Secondly, we excluded the Perceptual Loss. 
Without perceptual loss, structural preservation is downgraded and some noise started to showed up although still far less than Dreambooth since Hypnos also utilized inverse gaussian reconstruction loss.
Refer to suppl. material for details.

\subsubsection{Latent Discriminator Loss}
By removing the Latent Discriminator, the color consistency and background generation are severely impacted. 
This is reasonable as two-third of the training image scenes are just monotone colors, hence making the generated image have bad subject-scene blending.

\subsection{Limitations}
As shown on \cref{fig:Result}, Hypnos does not guarantee perfect image reconstruction nor 3D consistency.
Hypnos also never intended to be used on humans nor animals hence it may not be optimized for those tasks. 
There also seems to be a tradeoff between high foreground reconstruction and subject-scene blending capability.

\subsection{Societal Impact}
Our method enables small businesses, up to the well-established businesses, to easily gain access to a more reliable and flexible high-quality product image generation for marketing purposes compared to background image inpainting as the most common approach at the time of writing of this work.
On the other end of the spectrum, Hypnos can also potentially produce misleading images without any reality backing, which may lead to unintentional or even deliberate fraud or other malicious activities.

\section{Conclusion}

We propose Hypnos, a content-centric personalization diffusion finetuning technique that addresses high noise and foreground-background entanglement in T2I tasks, which are prevalent problems of its predecessors.
Given 3-5 image input samples, Hypnos consistently preserves the foreground's structure and color while enabling diverse background modification.
We also show that it is also possible to adjust the diversity of the generated image semantically, with inanimate object constrain, by utilizing our proposed supervision mechanisms. 
Furthermore, with the newly introduced hyperparameters, Hypnos offers a wider space of adjustment on the target distribution to support diverse use cases.
We believe that the development of Hypnos finetuning strategy paves the way for various foreground-focused downstream computer vision-based generative tasks.
\\
\\
\textbf{Acknowledgment}
This research project is supported by the grant of Penelitian Pemula Binus No: 069A/VRRTT/III/2024 (Universitas Bina Nusantara).

\bibliographystyle{splncs04}
\bibliography{main}

\begin{thebibliography}{10}
\providecommand{\url}[1]{\texttt{#1}}
\providecommand{\urlprefix}{URL }
\providecommand{\doi}[1]{https://doi.org/#1}

\bibitem{17-blattmann2023stable}
Blattmann, A., Dockhorn, T., Kulal, S., Mendelevitch, D., Kilian, M., Lorenz, D., Levi, Y., English, Z., Voleti, V., Letts, A., et~al.: Stable video diffusion: Scaling latent video diffusion models to large datasets. arXiv preprint arXiv:2311.15127  (2023)

\bibitem{30-caron2021emerging}
Caron, M., Touvron, H., Misra, I., J{\'e}gou, H., Mairal, J., Bojanowski, P., Joulin, A.: Emerging properties in self-supervised vision transformers. In: Proceedings of the IEEE/CVF international conference on computer vision. pp. 9650--9660 (2021)

\bibitem{21-chauhan2023brief}
Chauhan, V.K., Zhou, J., Lu, P., Molaei, S., Clifton, D.A.: A brief review of hypernetworks in deep learning. arXiv preprint arXiv:2306.06955  (2023)

\bibitem{10-BERT}
Devlin, J., Chang, M.W., Lee, K., Toutanova, K.: Bert: Pre-training of deep bidirectional transformers for language understanding. arXiv preprint arXiv:1810.04805  (2018)

\bibitem{28-dosovitskiy2021image}
Dosovitskiy, A., et~al.: An image is worth 16x16 words: Transformers for image recognition at scale  (2021)

\bibitem{2-TextualInversion}
Gal, R., Alaluf, Y., Atzmon, Y., Patashnik, O., Bermano, A.H., Chechik, G., Cohen-Or, D.: An image is worth one word: Personalizing text-to-image generation using textual inversion. arXiv preprint arXiv:2208.01618  (2022)

\bibitem{32-FID}
Heusel, M., Ramsauer, H., Unterthiner, T., Nessler, B., Hochreiter, S.: Gans trained by a two time-scale update rule converge to a local nash equilibrium. Advances in neural information processing systems  \textbf{30} (2017)

\bibitem{3-DDPM}
Ho, J., Jain, A., Abbeel, P.: Denoising diffusion probabilistic models. Advances in neural information processing systems  \textbf{33},  6840--6851 (2020)

\bibitem{19-hu2021lora}
Hu, E.J., Shen, Y., Wallis, P., Allen-Zhu, Z., Li, Y., Wang, S., Wang, L., Chen, W.: Lora: Low-rank adaptation of large language models. arXiv preprint arXiv:2106.09685  (2021)

\bibitem{14-Kingma_2019}
Kingma, D.P., Welling, M., et~al.: An introduction to variational autoencoders. Foundations and Trends{\textregistered} in Machine Learning  \textbf{12}(4),  307--392 (2019)

\bibitem{24-lee2022tracer}
Lee, M.S., Shin, W., Han, S.W.: Tracer: Extreme attention guided salient object tracing network (student abstract). In: Proceedings of the AAAI conference on artificial intelligence. vol.~36, pp. 12993--12994 (2022)

\bibitem{22-lu2024dreamcom}
Lu, L., Zhang, B., Niu, L.: Dreamcom: Finetuning text-guided inpainting model for image composition. arXiv preprint arXiv:2309.15508  (2023)

\bibitem{23-motamed2023lego}
Motamed, S., Paudel, D.P., Van~Gool, L.: Lego: Learning to disentangle and invert concepts beyond object appearance in text-to-image diffusion models. arXiv preprint arXiv:2311.13833  (2023)

\bibitem{9-GLIDE}
Nichol, A., Dhariwal, P., Ramesh, A., Shyam, P., Mishkin, P., McGrew, B., Sutskever, I., Chen, M.: Glide: Towards photorealistic image generation and editing with text-guided diffusion models. arXiv preprint arXiv:2112.10741  (2021)

\bibitem{5-ImprovedDDPM}
Nichol, A.Q., Dhariwal, P.: Improved denoising diffusion probabilistic models. In: International conference on machine learning. pp. 8162--8171. PMLR (2021)

\bibitem{15-podell2023sdxl}
Podell, D., English, Z., Lacey, K., Blattmann, A., Dockhorn, T., M{\"u}ller, J., Penna, J., Rombach, R.: Sdxl: Improving latent diffusion models for high-resolution image synthesis. arXiv preprint arXiv:2307.01952  (2023)

\bibitem{25-Qin_2020}
Qin, X., Zhang, Z., Huang, C., Dehghan, M., Zaiane, O.R., Jagersand, M.: U2-net: Going deeper with nested u-structure for salient object detection. Pattern recognition  \textbf{106},  107404 (2020)

\bibitem{8-radford2021learning}
Radford, A., Kim, J.W., Hallacy, C., Ramesh, A., Goh, G., Agarwal, S., Sastry, G., Askell, A., Mishkin, P., Clark, J., et~al.: Learning transferable visual models from natural language supervision. In: International conference on machine learning. pp. 8748--8763. PMLR (2021)

\bibitem{11-T5}
Raffel, C., Shazeer, N., Roberts, A., Lee, K., Narang, S., Matena, M., Zhou, Y., Li, W., Liu, P.J.: Exploring the limits of transfer learning with a unified text-to-text transformer. Journal of machine learning research  \textbf{21}(140),  1--67 (2020)

\bibitem{13-rombach2022highresolution}
Rombach, R., Blattmann, A., Lorenz, D., Esser, P., Ommer, B.: High-resolution image synthesis with latent diffusion models. In: Proceedings of the IEEE/CVF Conference on Computer Vision and Pattern Recognition (CVPR). pp. 10684--10695 (June 2022)

\bibitem{1-Dreambooth}
Ruiz, N., Li, Y., Jampani, V., Pritch, Y., Rubinstein, M., Aberman, K.: Dreambooth: Fine tuning text-to-image diffusion models for subject-driven generation. In: Proceedings of the IEEE/CVF conference on computer vision and pattern recognition. pp. 22500--22510 (2023)

\bibitem{20-ruiz2023hyperdreambooth}
Ruiz, N., Li, Y., Jampani, V., Wei, W., Hou, T., Pritch, Y., Wadhwa, N., Rubinstein, M., Aberman, K.: Hyperdreambooth: Hypernetworks for fast personalization of text-to-image models. In: Proceedings of the IEEE/CVF Conference on Computer Vision and Pattern Recognition. pp. 6527--6536 (2024)

\bibitem{12-saharia2022photorealistic}
Saharia, C., Chan, W., Saxena, S., Li, L., Whang, J., Denton, E.L., Ghasemipour, K., Gontijo~Lopes, R., Karagol~Ayan, B., Salimans, T., et~al.: Photorealistic text-to-image diffusion models with deep language understanding. Advances in neural information processing systems  \textbf{35},  36479--36494 (2022)

\bibitem{29-sauer2023adversarial}
Sauer, A., Lorenz, D., Blattmann, A., Rombach, R.: Adversarial diffusion distillation. arXiv preprint arXiv:2311.17042  (2023)

\bibitem{27-schwartz2023discriminative}
Schwartz, I., Sn{\ae}bjarnarson, V., Chefer, H., Belongie, S., Wolf, L., Benaim, S.: Discriminative class tokens for text-to-image diffusion models. In: Proceedings of the IEEE/CVF International Conference on Computer Vision. pp. 22725--22735 (2023)

\bibitem{16-shi2020improving}
Shi, Z., Zhou, X., Qiu, X., Zhu, X.: Improving image captioning with better use of captions. arXiv preprint arXiv:2006.11807  (2020)

\bibitem{31-sohldickstein2015deepunsupervisedlearningusing}
Sohl-Dickstein, J., Weiss, E., Maheswaranathan, N., Ganguli, S.: Deep unsupervised learning using nonequilibrium thermodynamics. In: International conference on machine learning. pp. 2256--2265. PMLR (2015)

\bibitem{4-DDIM}
Song, J., Meng, C., Ermon, S.: Denoising diffusion implicit models. arXiv preprint arXiv:2010.02502  (2020)

\bibitem{26-tan2020efficientnet}
Tan, M., Le, Q.: Efficientnet: Rethinking model scaling for convolutional neural networks. In: International conference on machine learning. pp. 6105--6114. PMLR (2019)

\bibitem{6-PFGM}
Xu, Y., Liu, Z., Tegmark, M., Jaakkola, T.: Poisson flow generative models. Advances in Neural Information Processing Systems  \textbf{35},  16782--16795 (2022)

\bibitem{7-PFGM++}
Xu, Y., Liu, Z., Tian, Y., Tong, S., Tegmark, M., Jaakkola, T.: Pfgm++: Unlocking the potential of physics-inspired generative models. In: International Conference on Machine Learning. pp. 38566--38591. PMLR (2023)

\bibitem{18-zhang2023adding}
Zhang, L., Rao, A., Agrawala, M.: Adding conditional control to text-to-image diffusion models. In: Proceedings of the IEEE/CVF International Conference on Computer Vision. pp. 3836--3847 (2023)

\bibitem{33-LPIPS}
Zhang, R., Isola, P., Efros, A.A., Shechtman, E., Wang, O.: The unreasonable effectiveness of deep features as a perceptual metric. In: Proceedings of the IEEE conference on computer vision and pattern recognition. pp. 586--595 (2018)

\end{thebibliography}

\clearpage
\appendix
\begin{table}[tb]
    \caption{Proportion of Latent Discriminator's Training set, the top three correspond to the real images set while the bottom five correspond to the fake images set}
    \label{tab:ld_proprtion}
    \centering
    \begin{tabular}{lllc}
    \hline
    Image Type  && \multicolumn{2}{c}{Proportion}      \\ \hline
    Instance Image                                          && \multicolumn{1}{l|}{30\%}    &      \\
    Colored   Background                                    && \multicolumn{1}{l|}{18\%}    & 50\% \\
    Colored   Background + resize                           && \multicolumn{1}{l|}{2\%}     &      \\ \hline
    Preservation/Class   Image                              && \multicolumn{1}{l|}{17,50\%} &      \\
    Negative   Foreground (Instance Image)                                   && \multicolumn{1}{l|}{9,75\%}  &      \\
    Masked   Foreground (Instance Image)                                     && \multicolumn{1}{l|}{3,25\%}  & 50\% \\
    Colored   Background + Negative Foreground (Instance Image)              && \multicolumn{1}{l|}{4,88\%}  &      \\
    Colored   Background + Masked Foreground (Instance Image)                && \multicolumn{1}{l|}{14,63\%} &      \\ \hline
    \end{tabular}
\end{table}

\section{Preliminary}
In this supplementary document, we provide implementation details to ensure reproducibility (written in \cref{sec:supp_imp_detail}), which includes in-depth strategy and intuition of our approach.
Within this section, we include the additional details on the proposed supervision mechanism (\cref{sec:supp_perc_loss}) and configurations on our model architecture (\cref{sec:supp_lat_disc}) according to the main paper's references.
Additionally, to elaborate further on our methodology, we present additional visualizations and a detailed discussion of the dataset and the inverse Gaussian function (written in \cref{sec:supp_data_images} and \cref{sec:supp_inv_gauss}). 
Finally, to offer a comprehensive overview of Hypnos, we include further ablation studies and an analysis on failure cases (written in \cref{sec:supp_abl_study} and \cref{sec:supp_failure_case}).

\section{Implementation Details}
\label{sec:supp_imp_detail}
To train the models we utilize L4 GPU with 22 GB VRAM, however we also had tested that V100 GPU with 16 GB VRAM is sufficient.

All of the models that are used in \cref{sec:result} are trained for 800 steps except for Textual Inversion, which we trained for 1500 steps.
We picked 1500 steps for Textual Inversion \cite{2-TextualInversion} because it took approximately the same training time as the 800 steps Dreambooth \cite{1-Dreambooth}.
These models are also trained on 8-bit Adam and quantized VAE.
Importantly, to ensure reproducibility, we have set all random seeds to 42.
Our code is written in the Pytorch framework.

\subsection{Perceptual Loss}
\label{sec:supp_perc_loss}

As mentioned on \cref{sec:loss}, to ensure foreground consistency, we opt to set larger weight for the shallow activations, which in detail are \textbf{0.35 for the second block activations}, \textbf{0.45 for the third block activations}, and \textbf{0.2 for the fourth block activations}. 
Note that this weight number is mostly arbitrary.
We advise adjusting the weight to adapt based on the given object and use cases.

\begin{figure}[t]
    \centering
    \includegraphics[width=1\linewidth]{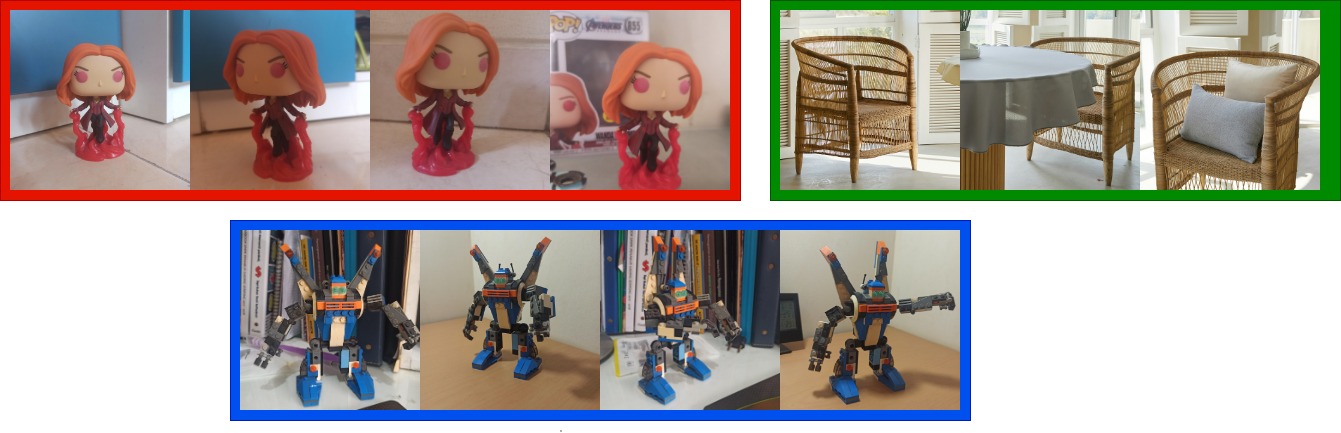}
    \caption{Instance Images, from left to right, top to bottom, {\color{red} Funko figurine}, {\color{green}Rattan chair}, and {\color{blue}Lego Robot}. }
    \label{fig:datset}
\end{figure}

\subsection{Latent Discriminator}
\label{sec:supp_lat_disc}

As seen on \cref{fig:ld_model}, the images are firstly encoded into latent space using VAE encoder ($\mathcal{E}$) and then fed through a classic inverted bottleneck convolutional layer, which yields a higher level feature extraction. 
The output then gets concatenated with the original latent image. 
It then split into 8x8 patches and summed with positional embeddings. 
The transformer consists of 3 blocks with 3 MLP heads for the [CLS] output embedding.

To prepare the Latent Discriminator with enough knowledge, we pretrained the model for 600 steps. 
We prepared a dataloader with diverse sampled image types as seen on \cref{tab:ld_proprtion}. 
These image types are utilized to ensure that the discriminator has a good understanding of foreground distinction, color, and structure. 
These image types are divided into real images and fake images.
The procedure to generate the real images is identical to the one used on the main training loop as mentioned on \cref{sec:image_aug}.
On the other hand, the fake image extraction procedure introduces a more diverse approach; the most intuitive method is to include preservation images as the fake samples; this addition is meant to enforce the model's understanding of structures and differentiate one object from another.
Besides using the preservation images, we also used the altered foreground version of the instance images.
The first modification is the usage of negative colors with the same background; this approach ensures the latent discriminator understands color consistency.
Second, the masked foreground strategy is applied to teach the model to avoid relying on the background region to distinguish real or fake images.
Note that we aim to ensure the discriminator has a clear focus on the foreground region. 
Hence, we opt to label images with foreground and background modification as fake and real data, respectively.

\subsection{Evaluation}
To evaluate metrics on both Prompt Invariant and Varying, we calculate the mean and standard deviations of each metric across 50 generated images. 
For each image that is generated by Hypnos, we compared it with each instance-images, as shown in \cref{fig:datset}.
For example, in a dataset that is comprised of 4 images, by this approach, the metrics will be calculated across 200 different evaluations ($4 \times 50$ evaluations).

\section{Dataset Images}
\label{sec:supp_data_images}
\cref{fig:datset} shows the instance images that we utilized for evaluation purposes on \cref{sec:result}. 
In the case of Funko figurine and Lego Robot, we use personal images taken using mobile phones to show that using amateur photos is already sufficient to make Hypnos generate visually pleasing image quality.
Note that poor lighting and shadows, as seen in \cref{fig:datset}, might still cause color distortions.

\section{Inverse Gaussian}
\label{sec:supp_inv_gauss}
\begin{figure}[t]
    \centering
      \begin{subfigure}{0.48\linewidth}
        \centering
        \includegraphics[width=.8\linewidth]{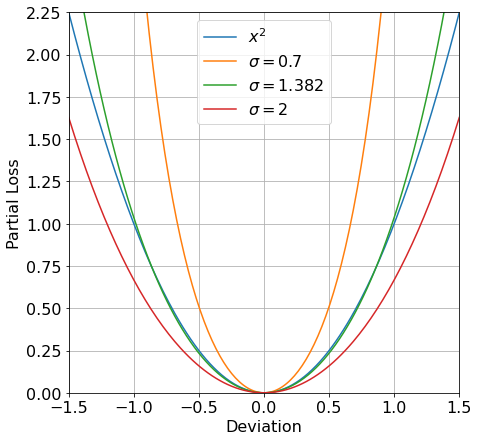}
        \caption{Inverse Gaussian comparisons}
        \label{fig:InvGaussGraph}
      \end{subfigure}
      \begin{subfigure}{0.47\linewidth}
        \centering
        \includegraphics[width=1\linewidth]{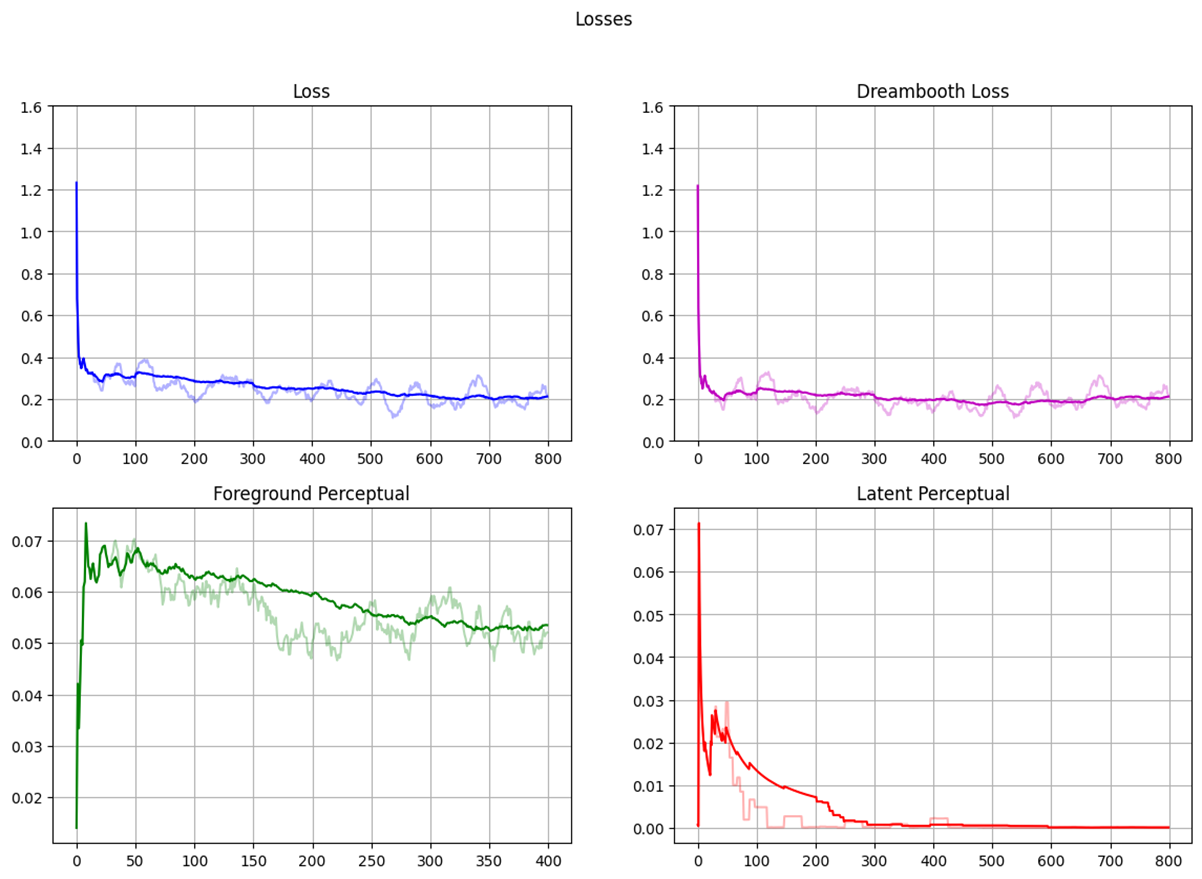}
        \caption{Learning Curve}
        \label{fig:curve}
      \end{subfigure}
    \caption{(a) shows varying $\sigma$ affects Inverse Gaussian function ($(1/{gaussian(x)})-1$) in partial loss space compared to quadratic MSE function, (b) shows the stability of learning curve.}
    \label{fig:invgauss}
\end{figure}

As seen in fig \cref{fig:InvGaussGraph}, employing \cref{eq:invertedgauss} allows for the flexible adjustment of the steepness of the curve. 
This loss is an exponential function; hence, it exhibits a far larger loss for high deviations compared to MSE. 
Usually, this approach is avoided because of the possible training instability.
Fortunately, Hypnos finetunes a pretrained network; therefore, the loss is often already very close to 0 and rarely exceeds 1, resulting in a stable training cycle as seen on the learning curve (\cref{fig:curve}).

\section{Supplementary Ablation Study}
\label{sec:supp_abl_study}
\subsection{Variation of Generated Images}
\begin{figure}[t]
    \centering
    \includegraphics[width=1\linewidth]{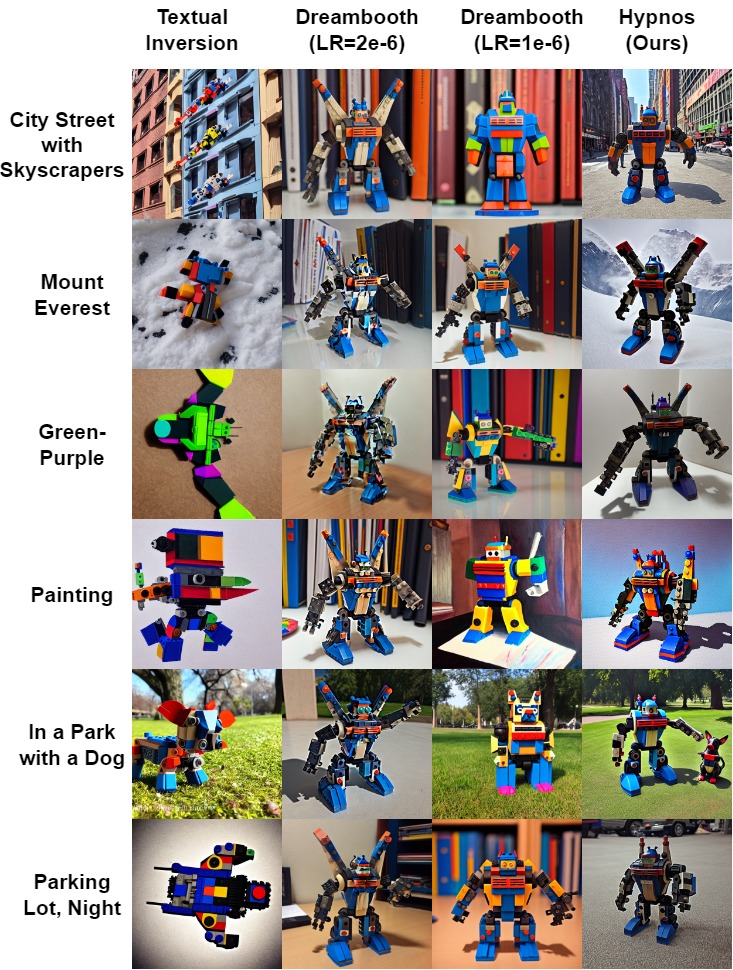}
    \caption{Lego Robot comparison (electronic screen viewing is advised)}
    \label{fig:robot_supp}
\end{figure}

\begin{figure}[t]
    \centering
    \includegraphics[width=1\linewidth]{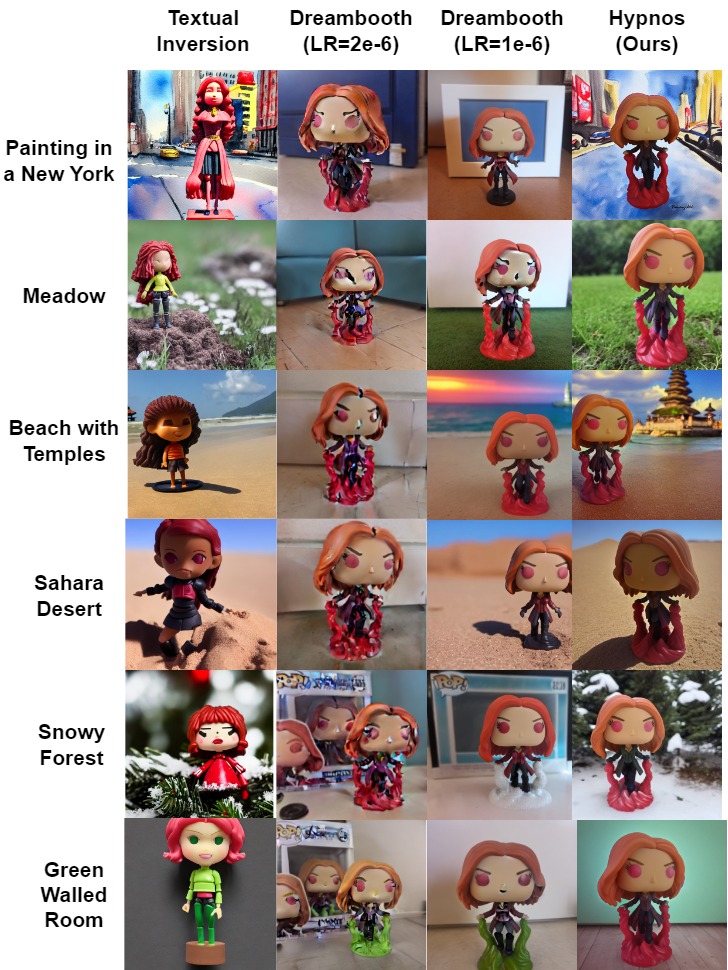}
    \caption{Funko Figurine comparison (electronic screen viewing is advised)}
    \label{fig:figurine_supp}
\end{figure}

\begin{figure}[t]
    \centering
    \includegraphics[width=1\linewidth]{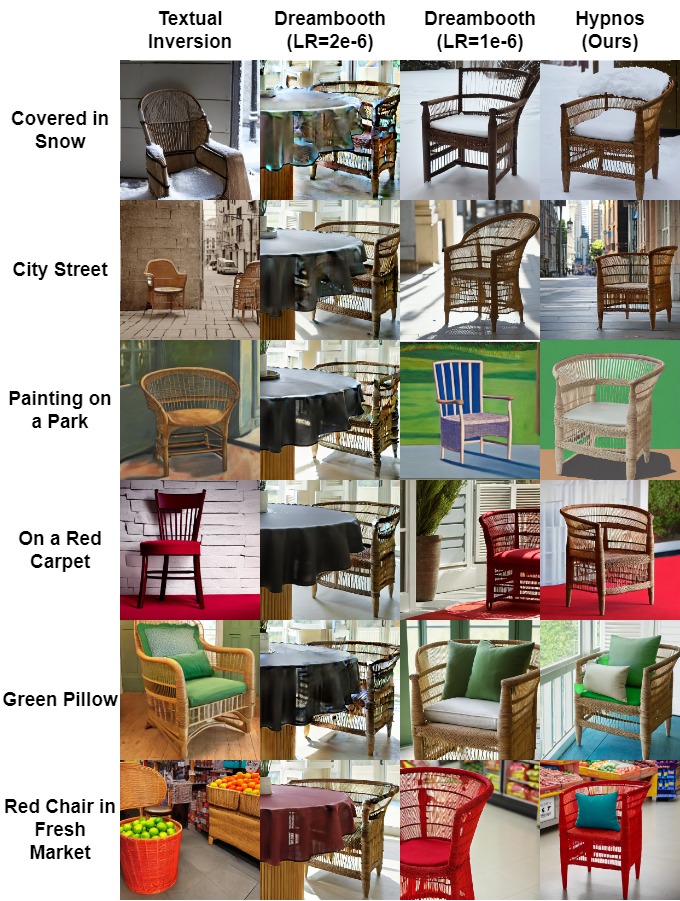}
    \caption{Rattan Chair comparison (electronic screen viewing is advised)}
    \label{fig:rattan_supp}
\end{figure}

Throughout our observations throughout the evaluation process on both qualitative and quantitative metrics, we observed that Dreambooth family techniques, which include our proposed method, have high variations in generated images.
We observed that even with the same trained model, by reevaluating the evaluation prochedure, it is possible to acquire a score more than one standard deviation away from the previous measurement. 
Qualitatively, Dreambooth often generates a high noised image, and on the other generated images, it is possible to stumble across neat images output as shown in \cref{fig:robot_supp}, \cref{fig:figurine_supp}, and \cref{fig:rattan_supp}. 
On the other hand, Hypnos exhibit a more consistent image quality throughout the generation process.
Based on this understanding, we recommend sampling several images and conducting a manual assessment prior to their utilization in downstream tasks.

\subsection{Perceptual Loss and Latent Discriminator Loss}
\begin{figure}[t]
    \centering
    \includegraphics[width=1\linewidth]{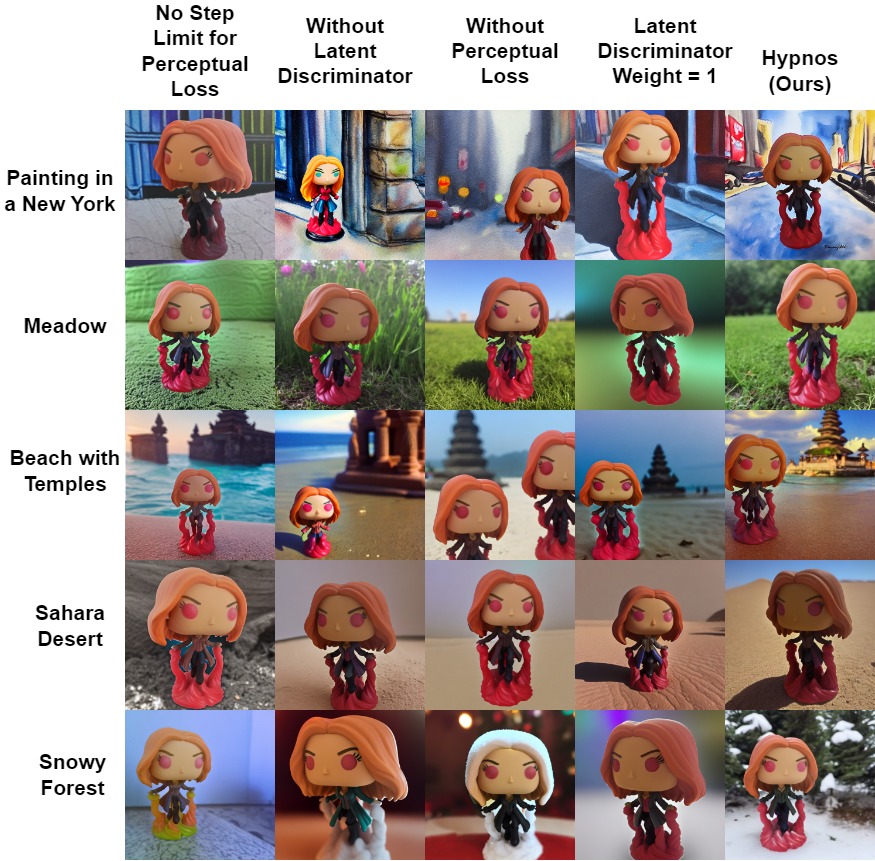}
    \caption{Funko Figurine ablation study (electronic screen viewing is advised)}
    \label{fig:ablation_supp}
\end{figure}

\cref{fig:ablation_supp} shows the visualization of ablation study experiments mentioned on \cref{sec:ablation}.
As shown in figure \cref{fig:ablation_supp}, Perceptual Loss and Latent Discriminator are complementary losses.
In other words, the absence of one results in a deterioration of the quality of the generated images.
This is reasonable as both loss works in different spaces and perspectives as explained on \cref{sec:loss}.

Both losses are also unsuitable to replace reconstruction loss to guide the optimization process (shown by the increase of weight on the first and fourth columns of \cref{fig:ablation_supp}).
This is supported by the solid mathematical derivation of reconstruction loss and the nature of Perceptual Loss and Latent Discriminator Loss that may vary based on the reliability of the corresponding neural network models (e.g., EfficientNetB1, ViT).

\begin{figure}[t]
    \centering
    \includegraphics[width=1\linewidth]{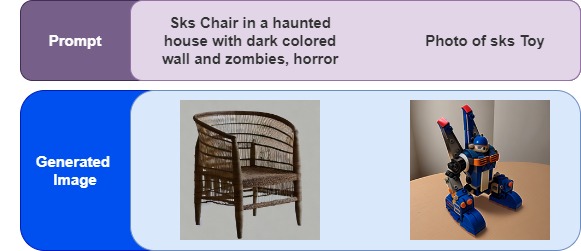}
    \caption{Examples of Failure Cases. (left) shows an image that fails to adapt to the given prompt, (right) shows an altered object with one arm missing}
    \label{fig:failure}
\end{figure}

\section{Failure Cases Analysis}
\label{sec:supp_failure_case}
Throughout our evaluation, we also assessed some failure cases to better understand the behavior and limitations of Hypnos.
\cref{fig:failure} shows examples of failure cases generated by Hypnos.
The first shows a blank background when given a complex prompt.
We suspect this is partly due to the base model limitation, we justify this by confirming that the base model still fails to generate the same prompt even with a normal chair.
We also observed that Hypnos tends to be more conservative in generating foreground variations with a trade off in foreground consistency.
This phenomenon can be minimized by decreasing the changed background ratio and decreasing the perceptual and latent discriminator strength.

We observed that the provided preservation dataset also influenced background and scene generation capability.
Supplying several preservation images with a specific background can enable the model to generate similar backgrounds.
We anticipate that this effect may be less significant for larger base models, as they typically possess broader text-to-image (T2I) capabilities. 
However, further justification and experiments on this matter are beyond the scope of this work and may be addressed in future research.

The second image from the left (\cref{fig:failure}) showed a less similar foreground with the absence of one forearm.
This is due to the limitation of the background remover model, which accidentally removes the arm.
This issue can be overcome by excluding the problematic image, reducing the change in background ratio, or even replacing the background remover model.
On the other hand, the underfitting can be explained by the complexity of the object. 
Hence, this can be trivially overcome by increasing the learning rate. 
Hypnos shows to be compatible with a high learning rate as it is still able to generate noiseless images.

In other cases where generating a better foreground is more favorable than training efficiency, we advise incorporating a stronger and deeper discriminator as it may increase the image quality.
Despite the limitations above, Hypnos proves that the proposed finetuning strategy can accommodate the current prominent T2I model in handling foreground-focused generative task in a straightforward manner.

\end{document}